\def\year{2022}\relax
\documentclass[letterpaper]{article} 
\usepackage{aaai22}  
\usepackage{times}  
\usepackage{helvet}  
\usepackage{courier}  
\usepackage[hyphens]{url}  
\usepackage{graphicx} 
\usepackage{natbib}  
\usepackage{caption} 

\usepackage{amsmath}
\usepackage{microtype}
\usepackage{multirow}
\usepackage{soul}
\usepackage{url}
\usepackage{graphicx}
\usepackage{epstopdf}
\usepackage{subfigure}

\usepackage{amsthm}
\usepackage{booktabs}
\usepackage{algorithm}
\usepackage{algorithmic}
\usepackage{xspace}

\usepackage{enumitem}
\setlist[itemize]{leftmargin=*}

\usepackage{eqparbox}

\DeclareCaptionStyle{ruled}{labelfont=normalfont,labelsep=colon,strut=off} 
\usepackage{algorithm}
\usepackage{algorithmic}

\usepackage{newfloat}
\usepackage{listings}
\floatstyle{ruled}
\newfloat{listing}{tb}{lst}{}
\floatname{listing}{Listing}

\usepackage{color}

\newcommand{\modelname}{CycTEA\xspace}
\title{Ensemble Semi-supervised Entity Alignment via Cycle-teaching}
\author{
    Kexuan Xin\textsuperscript{\rm 1},
    Zequn Sun\textsuperscript{\rm 2},
    Wen Hua\textsuperscript{\rm 1}\footnotemark[1],
    Bing Liu\textsuperscript{\rm 1},
    Wei Hu\textsuperscript{\rm 2},
    Jianfeng Qu\textsuperscript{\rm 3}\thanks{Corresponding authors},
    Xiaofang Zhou\textsuperscript{\rm 4}
}
\affiliations{
    \textsuperscript{\rm 1}The University of Queensland, Brisbane, QLD 4072, Australia\\
    \textsuperscript{\rm 2}State Key Laboratory for Novel Software Technology, Nanjing University, China\\
    \textsuperscript{\rm 3}Soochow University, Suzhou, Jiangsu 215006, China\\
    \textsuperscript{\rm 4}Hong Kong University of Science and Technology, Kowloon, Hong Kong\\
    \{uqkxin, w.hua, bing.liu\}@uq.edu.au, zqsun.nju@gmail.com, whu@nju.edu.cn, jfqu@suda.edu.cn, zxf@cse.ust.hk
}

\usepackage{bibentry}

\begin{document}

\maketitle

\begin{abstract}
Entity alignment is to find identical entities in different knowledge graphs. Although embedding-based entity alignment has recently achieved remarkable progress, training data insufficiency remains a critical challenge. Conventional semi-supervised methods also suffer from the incorrect entity alignment in newly proposed training data. To resolve these issues, we design an iterative cycle-teaching framework for semi-supervised entity alignment. The key idea is to train multiple entity alignment models (called aligners) simultaneously and let each aligner iteratively teach its successor the proposed new entity alignment. We propose a diversity-aware alignment selection method to choose reliable entity alignment for each aligner. We also design a conflict resolution mechanism to resolve the alignment conflict when combining the new alignment of an aligner and that from its teacher. Besides, considering the influence of cycle-teaching order, we elaborately design a strategy to arrange the optimal order that can maximize the overall performance of multiple aligners. The cycle-teaching process can break the limitations of each model's learning capability and reduce the noise in new training data, leading to improved performance. Extensive experiments on benchmark datasets demonstrate the effectiveness of the proposed cycle-teaching framework, which significantly outperforms the state-of-the-art models when the training data is insufficient and the new entity alignment has much noise.
\end{abstract}

\section{Introduction}
Entity alignment seeks to find identical entities of different KGs that refer to the same real-world object. 
Recently, embedding-based entity alignment approaches have achieved great progress \cite{sun2017cross,sun2018bootstrapping,cao2019multi,wu2019relation,sun2020knowledge,zeng2020degree,mao2020relational}.
One prime advantage of embedding techniques lies in relieving the heavy reliance on hand-craft features or rules. 
KG embeddings have also demonstrated their great strength to tackle the symbolic heterogeneity of different KGs \cite{wang2017knowledge}.
Especially for entity alignment, embedding-based approaches capture the similarity of entities in vector space. 
But these approaches highly rely on sufficient training data (i.e., seed entity alignment) to bridge different KG embedding spaces for alignment learning.
The training data insufficiency issue in real-world scenarios \cite{chen2017multilingual} prevents embedding-based approaches from effectively capturing entity similarities across different KGs.

The semi-supervised approach is an effective solution to the above issue. 
It iteratively proposes new ``reliable'' entity alignment to augment training data \cite{zhu2017iterative,sun2018bootstrapping,chen2018co}.
There are several shortcomings of existing semi-supervised approaches,
and not all semi-supervised strategies can bring improved performance and stability to entity alignment in real scenarios \cite{sun2020benchmarking}.
For example, although the popular self-training approach BootEA \cite{sun2018bootstrapping} can alleviate the error-accumulation issue via a heuristic alignment editing method,
the learning capability of its entity alignment model and the alignment selection bias still limit its performance.
The co-training approach KDCoE \cite{chen2018co} incorporates literal descriptions as side information to complement the structure view of entities. 
However, it requires a high complementary of the two independent feature views and specific prior knowledge for feature selection. 
It usually fails to bring improvement due to the limited availability of descriptions.

In summary, the shortcomings of existing semi-supervised entity alignment approaches lie in the following aspects. 
(\romannumeral1) \textbf{Noisy alignment accumulation}.
This is the critical challenge for semi-supervised approaches that the newly proposed entity alignment inevitably contains much noisy data. 
Iteratively accumulating new entity alignment as training data is also error-propagation. 
The incorrect entity alignment can spread to the following iterations with adverse effects on final performance.
(\romannumeral2) \textbf{Biased alignment selection}.
The semi-supervised approach usually proposes the predicted high-confidence alignment to bootstrap itself, 
and such alignment will receive more training and higher confidence in the next iterations. 
The approach will then be more inclined to propose the same more and more ``reliable'' alignment, leading to a biased alignment selection. 
The performance improvement brought by retraining the entity alignment model with what it already knows is limited.
(\romannumeral3) \textbf{Performance bottleneck of the aligner}. 
Although the embedding-based entity alignment model (called aligner) can receive better optimization with more training data during semi-supervised learning, 
the aligner has its performance bottleneck due to its limited expressiveness on embedding learning.
This can be reflected by the variable performance of the entity alignment approach when employing different embedding models, e.g., TransE \cite{bordes2013translating} and GCN~\cite{kipf2017gcn}.

To address the above shortcomings,
we elaborately design a novel \textbf{Cyc}le-\textbf{T}eaching framework for \textbf{E}ntity \textbf{A}lignment, named \textbf{\modelname},
which enables multiple entity alignment models (called aligners) to teach each other. 
\modelname lets each aligner teach its selected new entity alignment to its subsequent aligner for robust semi-supervised training.
The subsequent aligner can filter noisy alignment via alignment conflict resolution and get more reliable entity alignment to augment training data. 
The motivation behind our work is that, 
as different aligners have different alignment capacities, 
the selected new entity alignment of an aligner can benefit other aligners and help them filter the noisy alignment introduced by the biased alignment selection \cite{han2018coteach}.

Cycle-teaching possesses some critical advantages over the traditional ensemble semi-supervised method, e.g., Tri-Training \cite{Zhou05tritraining} that integrates three models in the ``majority-teach-minority'' way (i.e., majority vote).
First, cycle-teaching can help each aligner break its performance bottleneck. It can produce more diverse and complementary entity alignment since the aligners have different capacities and are trained based on their own training data. 
Taught by the new ``knowledge'' from others, each aligner can overcome the ceiling of entity alignment performance.
Second, cycle-teaching can reduce the risk of noise training data.
In cycle-teaching, as different aligners have different learning abilities, they can filter different types of incorrect entity alignment by the proposed diversity-aware alignment selection and conflict resolution. 
The error flows from one aligner to its successor can be reduced during the iterations.
Third, cycle-teaching can be easily extended to multiple aligners (more than three and also even number of aligners).
It can avoid the problem that multiple models fail to reach an agreement by majority vote.
Our contributions are summarized as follows:
\begin{itemize}
\item We propose a novel semi-supervised learning framework, i.e., cycle-teaching, for entity alignment. 
It seeks to build a strong and robust entity alignment approach by integrating multiple simple aligners. 
It does not require sufficient feature views of entities or high performance of each aligner, 
and is able to achieve better generalization ability.
\item To guarantee the quality of new entity alignment, 
we propose a diversity-aware alignment selection method and resolve alignment conflict by re-matching. 
We determine the cycle-teaching order based on the complementarity and performance difference of neighboring aligners.
The cycle-teaching paradigm helps the multiple aligners combat the noise alignment during iterative training. 
For each aligner, its new entity alignment combined with the new knowledge learned from others can bring significant performance gain.
\item We show that conventional semi-supervised methods, e.g., self-training and co-training can be regarded as the special cases of cycle-teaching. 
The advantages of cycle-teaching lie in reducing noise alignment accumulation and markedly boosting each aligner by teaching it unknown alignment.
\item Our framework can integrate any entity alignment models, including relation-based models such as AlignE \cite{sun2018bootstrapping}, RSN4EA \cite{guo2018recurrent} and AliNet \cite{sun2020knowledge}.
Extensive experiments on the benchmark entity alignment datasets OpenEA \cite{sun2020benchmarking} demonstrate the effectiveness of our framework.
\end{itemize}

\section{Related Work}
\label{sec:related_work}

\smallskip
\noindent\textbf{Structure-based Entity Alignment.}
The assumption for structure-based entity alignment is that similar entities should have similar relational structures. Early studies such as MTransE \cite{chen2017multilingual}, AlignE \cite{sun2018bootstrapping}, SEA \cite{pei2019semi} exploit TransE \cite{bordes2013translating} as the base embedding model for relational structure learning. To capture entity alignment across different KGs, the two KGs are merged as one graph for joint embedding or separately embedded along with a linear mapping. Recent studies such as GCN-Align \cite{wang2018cross}, AliNet \cite{sun2020knowledge} and others \cite{cao2019multi,zhu2019neighborhood,li2019semi,fey2020deep,ye2019vectorized,xu2019cross} design various graph neural networks (GNNs) for neighborhood structure learning and alignment learning. Some approaches that exploit long-term relational dependency of entities like IPTransE \cite{zhu2017iterative} and RSN4EA \cite{guo2018recurrent} have also achieved great progress.

\smallskip
\noindent\textbf{Attribute-enhanced Entity Alignment.}
Other approaches enhance entity alignment by learning from side information such as attribute correlation \cite{sun2017cross}, attribute values \cite{trisedya2019entity,zhang2019multi}, entity names \cite{wu2019relation,wu2019jointly,wu2020neighborhood,liu2020exploring} and distant supervision information from pre-trained language models \cite{yang2019aligning,tang2020bertint}. Although achieving high performance, one major problem of these models lies in their limited generalizability since the side information is not always available in different KGs.

\smallskip
\noindent\textbf{Semi-supervised Entity Alignment.}
As seed entity alignment is usually limited in real scenarios, some approaches explore to label new alignment to augment training data iteratively. IPTransE \cite{zhu2017iterative} conducts self-training to propose new alignment. However, it fails to achieve satisfying performance because it accumulates much noise data during iterations. Some work \cite{chen2018co,COTSAE} uses a co-training mechanism to propagate new alignment from two orthogonal views (e.g., relational structures and attributes). However, the improvement is also limited because some entities do not have attributes. BootEA \cite{sun2018bootstrapping} implements a heuristic editing method to mitigate the error-propagation issue, bringing significant improvement. However, the new seed selection is also limited by the model performance. When the accumulated new pairs can be aligned successfully by the embedding module itself, the improvement would be smaller. Therefore, we aim to design an approach that iteratively labels reliable entity alignment as training data and accumulates the new entity alignment that one model can hardly find by itself based on cycle-teaching.

\section{Embedding-based Entity Alignment}
\label{sec:background}
We define a KG as a 3-tuple, i.e., $\mathcal{K}=(\mathcal{E}, \mathcal{R}, \mathcal{T})$. $\mathcal{E}$ and $\mathcal{R}$ denote entity and relation sets, respectively.
$\mathcal{T} \subseteq \mathcal{E}\times \mathcal{R}\times\mathcal{E}$ is the set of relational triples.
Following \cite{sun2018bootstrapping}, we consider entity alignment between a source KG $\mathcal{K}_1=(\mathcal{E}_1, \mathcal{R}_1,\mathcal{T}_1)$ and a target one $\mathcal{K}_2=(\mathcal{E}_2,\mathcal{R}_2,\mathcal{T}_2)$. Given a small set of seed entity alignment $\mathcal{A}_{train}=\{(e_1, e_2) \in \mathcal{E}_1\times\mathcal{E}_2\|e_1\equiv e_2\}$ as training data, the task seeks to find the remaining entity alignment. For embedding-based approaches, the typical inference process is via the nearest neighbor search in the embedding space, i.e., given an aligned entity pair $(x, y)$, embedding-based approaches seek to hold
\begin{equation} 
\label{eq:inference}
\mathbf{y}=\arg \max_{y'\in \mathcal{E}_2} \pi(\mathbf{x},\mathbf{y}'),
\end{equation}
where $\pi(\mathbf{x},\mathbf{y})$ is a similarity measure to serve as the alignment confidence of entities and we use cosine in the paper.
Hereafter, we use bold-faced letters to denote embeddings, e.g, $\mathbf{x}$ and $\mathbf{y}$ are the embeddings of entities $x$ and $y$, respectively. To achieve the goal as in Eq. (\ref{eq:inference}), an entity alignment framework usually employs two basic modules: knowledge embedding and alignment learning \cite{sun2020benchmarking}.

\subsection{Knowledge Embedding}
\label{sec: kge}
This module seeks to learn an embedding function $f$ to map an entity to its embedding, i.e., $f(x) = \mathbf{x}.$ TransE \cite{bordes2013translating}, RSN4EA~\cite{guo2018recurrent} and GNN \cite{kipf2017gcn} are three popular KG embedding techniques. In TransE, the embeddings are learned by minimizing a energy function over each triple $(h,r,t)$:
\begin{equation} 
\label{eq:transe}
\min_f \sum_{(h,r,t)\in \mathcal{T}_1 \cup \mathcal{T}_2} \| f(h) + f(r) - f(t) \|,
\end{equation}
where $\|\cdot \|$ denotes $L_1$ or $L_2$ vector norm. KG embeddings can be learned by jointly optimizing the TransE's objective and the alignment learning objective (in the next section). RSN4EA~\cite{guo2018recurrent} proposes a recurrent skip mechanism to capture the long-term semantic information. It uses a biased random walk to generate relation paths such as $\left(x_{1}, x_{2}, \dots, x_{T}\right)$ with entities and relations in an alternating order. It encodes the paths as the 
output hidden state in RNN, i.e., $\mathbf{o}_{i}=\tanh \left(\mathbf{W}_1 \mathbf{o}_{i-1}+\mathbf{W}_2 \mathbf{x}_{i}+\mathbf{b}\right)$ at step $i$, where $\mathbf{W}_1$, $\mathbf{W}_2$ are the weight matrices and $\mathbf{b}$ is the bias.
The skipping connection is defined as follows to enhance the semantic representations between entities and relations:
\begin{equation}\label{eq:RSN}
\mathbf{o}_{i}^{\prime}=\left\{
 \begin{array}{ll}{
 \mathbf{o}_{i}} & {x_{i} \in \mathcal{E}} \\ 
 {\mathbf{S}_{1} \mathbf{o}_{i}+\mathbf{S}_{2} \mathbf{x}_{i-1}} & {x_{i} \in \mathcal{R}}
 \end{array}\right.,
\end{equation}
where $\mathbf{S}_{1}$ and $\mathbf{S}_{2}$ are weight matrices for entities and relations, respectively. For GNN, $f$ is an aggregation function to combine the representations of central entity and neighbors:
\begin{equation} 
\label{eq:gnn}
f(e) = \text{comb}(\mathbf{e}, \text{agg}(\mathcal{N}_e)),
\end{equation}
where $\mathcal{N}_e$ are the embeddings of entity $e$'s neighbors. Different aggregation strategies lead to different GNN variants. GNNs output entity representations for alignment learning. 

All existing knowledge embedding techniques can be applied to our cycle-teaching framework. Specifically, TransE (e.g., AlignE \cite{sun2018bootstrapping}) captures the local semantics from relation triples, GNN (e.g., AliNet \cite{sun2020knowledge}) models the global structure of KGs, and RSN4EA~\cite{guo2018recurrent} leans the long-term semantic knowledge. 
In addition, other side information can also be considered by incorporating the attribute-enhanced aligners into the cycle-teaching framework, which is left for future work.

\subsection{Alignment Learning}
To capture the alignment information, some models directly maximize the embedding similarities of pre-aligned entities, whose objective can be formulated as follows:
\begin{equation} 
\label{eq:align}
\max \frac{1}{|\mathcal{A}_{train}|} \sum_{(x,y)\in \mathcal{A}_{train}}\pi(\mathbf{x},\mathbf{y}).
\end{equation}
Augmenting training data is our focus in this paper. 


\begin{figure*}[!t]
    \centering
    \includegraphics[width=0.75\textwidth]{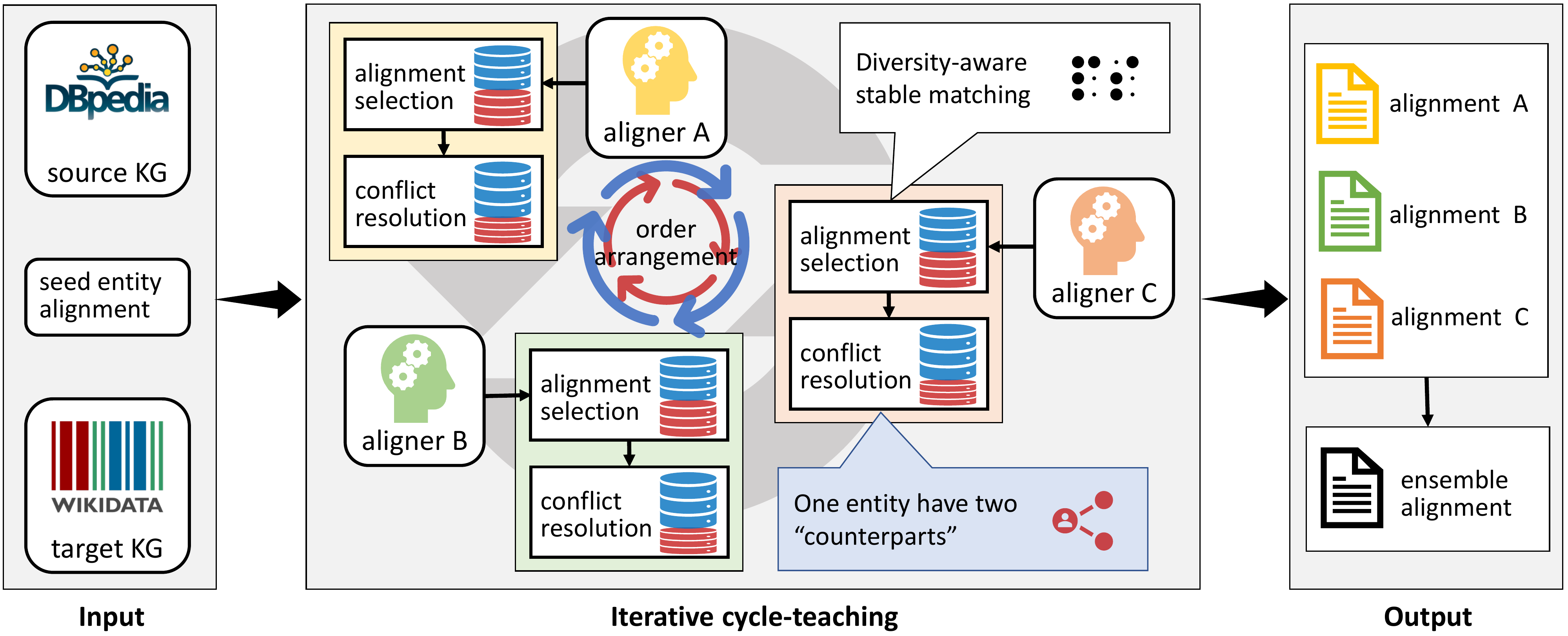}
    \caption{\label{fig:framework}Cycle-teaching framework for entity alignment.}
\end{figure*}

\section{Cycle Teaching for Entity Alignment}
\label{sec:cyctea}

Figure \ref{fig:framework} illustrates the cycle-teaching framework.
At each iteration,
if training has not been terminated, 
our framework would automatically compute an optimal cycle-teaching order (Sect. \ref{sect:order}).
Each aligner proposes reliable entity alignment pairs (Sect. \ref{sec: alignment selection}) and then transmits them to the successor. 
The successor aligner combines its own new entity alignment and the received ones via conflict resolution (Sect. \ref{sec:conflict_rematching}).
Then the resolved new entity alignment pairs are added into the training data set to train the aligner further, such that this aligner can be taught by the new entity alignment that it cannot find. 
Its peer can attenuate the effect of some noisy entity alignment from one model. 
When the training ends, the framework combines the results from all aligners to calculate the alignment ranking list for each source entity (Sect. \ref{sect:retrieval}).

\subsection{Cycle-teaching Order Arrangement}
\label{sect:order}

There are multiple aligners in \modelname.
Intuitively, the adjacent aligners should have higher complementarity such that the successor can receive more reliable alignment beyond its capacity. Moreover, it is better to let the aligner with higher performance teach weaker aligners so that the student aligner (successor) can be promoted by the more excellent teacher aligner (predecessor).
To this end, we formalize our order arrangement problem as a Travelling Salesman Problem (TSP). 
We first build a directed complete graph where each aligner works as a node, and the edge weight reveals how beneficial to connect the two nodes. Then, the task is to return the route starting from an aligner while ending with the same one and has the highest sum of edge weights. The resulted route indicates the order arrangement. The most important thing is to define the edge weight and we hereby consider two critical factors.
The first factor is the complementarity of the alignment selection from $M_i$ to $M_j$:
\begin{equation}
    f_{\text{com}}(M_i, M_j) = |(\mathcal{A}_i - \mathcal{A}_i \cap \mathcal{A}_j)|/|(\mathcal{A}_j)|,
\end{equation}
where $\mathcal{A}_i$ and $\mathcal{A}_j$ denote the new reliable alignment sets of $M_i$ and $M_j$ at current iteration, respectively.
It is noted that $f_{\text{com}}(M_i,M_j) \neq f_{\text{com}}(M_j, M_i)$ as we measure the complementarity feature in an asymmetric way, which reflects the new alignment brought by aligner $M_i$ to $M_j$.
We also want the stronger aligner to teach the weaker one. 
Therefore, we define the weight of performance between aligners $M_i$ and $M_j$ as the current Hits@1 difference on valid dataset:
\begin{equation}
    f_{\text{per}}(M_i, M_j) = \exp(\text{valid}(M_i) - \text{valid}(M_j)),
\end{equation}
Note that $f_{\text{per}}(M_i,M_j) \neq f_{\text{per}}(M_j,M_i)$ since the subtraction operation is not symmetric.
The final edge weight from aligner $M_i$ to $M_j$ is the combination of the two factors:
\begin{equation}
    w(M_i,M_j) = f_{\text{com}}(M_i, M_j) + \epsilon * f_{\text{per}}(M_i, M_j),
\end{equation}
where $\epsilon$ is the combination weight. After calculating each edge weight in the aligner graph, we aim to find an optimal path to traverse the whole graph covering the maximized edge weights. This TSP problem is NP-hard as there are totally $(k-1)!$ choices of paths). We can utilize existing TSP approximate algorithms to derive sub-optimal solution when the number of aligners is very large. While in practice, we can enumerate all paths and pick the optimal one since the number of aligners is usually small.

\subsection{Diversity-aware Alignment Selection}
\label{sec: alignment selection}
To pick out reliable entity alignment as new training data,
we propose diversity-aware alignment selection,
which considers both the embedding similarity and match diversity.

\smallskip
\noindent\textbf{Match Diversity.} 
Entity alignment is a $1:1$ matching task,
i.e., a source entity can be matched with at most one target entity, and vice versa \cite{suchanek2011paris}.
We expect a source entity to have very high embedding similarity with its counterpart in the target KG.
Existing methods use nearest neighbor (NN) search to retrieve entity alignment,
while ignoring the similarity distribution over other entities.
In contrast, we consider the match diversity \cite{mcd_www16}, which measures how much a predicted alignment pair $(x, y)$ deviates from other competing pairs like $(x', y)$ and $(x, y')$.
We compute the average similarity of all competing pairs as:
\begin{equation}\scriptsize
    \mu(x, y) = \frac{1}{(\|\mathcal{N}_{x}\| + \|\mathcal{N}_{y}\|-1)} \Big(\sum_{y'\in \mathcal{N}_{x}} \pi(x,y') + \sum_{x'\in \mathcal{N}_{y}}, \pi(x',y)\Big),
\end{equation}
where $\mathcal{N}_{x}$ denotes the set of all the candidate target entities for the source entity $x$ (including $y$), 
and $\mathcal{N}_{y}$ is the set of all candidate source entities for the target entity $y$ (including $x$). 
Given the average similarity, we define the match diversity of $(x, y)$ as the difference between its similarity and the average:
\begin{equation}
    \tau(x, y) = \pi(x,y) - \mu(x, y).
\end{equation}
We expect the correct entity alignment pair has a high match diversity, 
which indicates that the pairs have a high embedding similarity while its competing pairs have low similarity.

\smallskip
\noindent\textbf{Alignment Selection via Stable Matching.}
We use match diversity as alignment confidence to select new alignment.
To guarantee the quality of the selected entity alignment and satisfy the constraint of $1:1$ matching,
we model alignment selection as a stable matching problem (SMP).
We generate a sorted candidate list for each entity based on the alignment confidence. The SMP can be solved by the Gale–Shapley algorithm, which can produce a stable matching for all entities in time $O(n^2)$ where $n$ is the number of source entities.

\subsection{Conflict Resolution via Re-matching}
\label{sec:conflict_rematching}
For each aligner, it may have conflict selection between its predecessor and itself. For example, source entity $x$ may have two different ``counterparts'' $y_1$ and $y_2$ predicted by itself and its predecessor, respectively. 
Our solution is to let the two aligners cooperate to resolve the alignment conflicts.

For the selected reliable entity alignment of two aligners, $\mathcal{A}_1$ and $\mathcal{A}_2$, assuming that the conflict alignment set is $\mathcal{C}$, we collect the entities appearing in $\mathcal{C}$ to conduct a re-matching process. 
Specifically, for the involved entities from the source KG $\mathcal{X} = \{x|(x, y_1, y_2) \in \mathcal{C}, x \in \mathcal{E}_1, y_1,y_2\in \mathcal{E}_2\}$, 
and the entities that have not been matched from the target KG $\mathcal{Y} = \{y|y \in \mathcal{E}_2, y \not \in \mathcal{A}_1 \cup \mathcal{A}_2 \}$,
we utilize them to build a bipartite graph with weights and conduct the alignment selection in \ref{sec: alignment selection} to select more reliable alignment pairs. 
Considering the conflict pairs are difficult, 
we combine the similarity of two aligners, to serve as bipartite graph edge weights:
\begin{equation}
    \pi(x, y) = \alpha \cdot \pi_1(x, y) + (1 - \alpha) \cdot \pi_2(x, y),
\end{equation}
where $\alpha = \text{valid}_1/(\text{valid}_1 + \text{valid}_2)$ is a balance weight based on the two aligners' validation performance (Hits@1).

Compared with other possible conflict resolution strategies, such as majority vote and ensemble training, our method has the following advantages. The majority vote is limited to odd numbers of aligners, and it does not consider the similarity distribution, so the final selection is limited to the choices of aligners. 
For ensemble training, it outputs the same selection for all the aligners, and these aligners will become increasingly similar as the training process continues, resulting in lower robustness of alignment noise. In contrast, our re-matching strategy considers similarity distribution to repair incorrect alignment pairs, and the cycle propagation prevents all aligners to rapidly become similar. 

\subsection{Ensemble Entity Alignment Retrieval}
\label{sect:retrieval}

To benefit from all aligners, we combine their embedding similarity to generate final alignment result. We firstly assign weights $\{\alpha_1, \alpha_2, \dots, \alpha_k\} $ to aligners $\{M_1,M_2,\dots,M_k\}$ based on their Hits@1 on validation data:
\begin{equation}
    \alpha_i = \frac{\text{valid}(M_i)}{\sum_{j\in \{1,\dots,k\}}\text{valid}(M_j)}.
\end{equation}
Then, the final similarity between entities is defined as the weighted sum of the similarity of each aligner:
\begin{equation}
    \pi(x, y) = \sum_{i\in \{1,\dots,k\}}\alpha_i \cdot \pi_i(x, y).
\end{equation}
Given the ensemble entity similarities, we obtain the candidate counterpart list for each source entity by the NN search.

\begin{figure}
    \centering
    \includegraphics[width=0.75\linewidth]{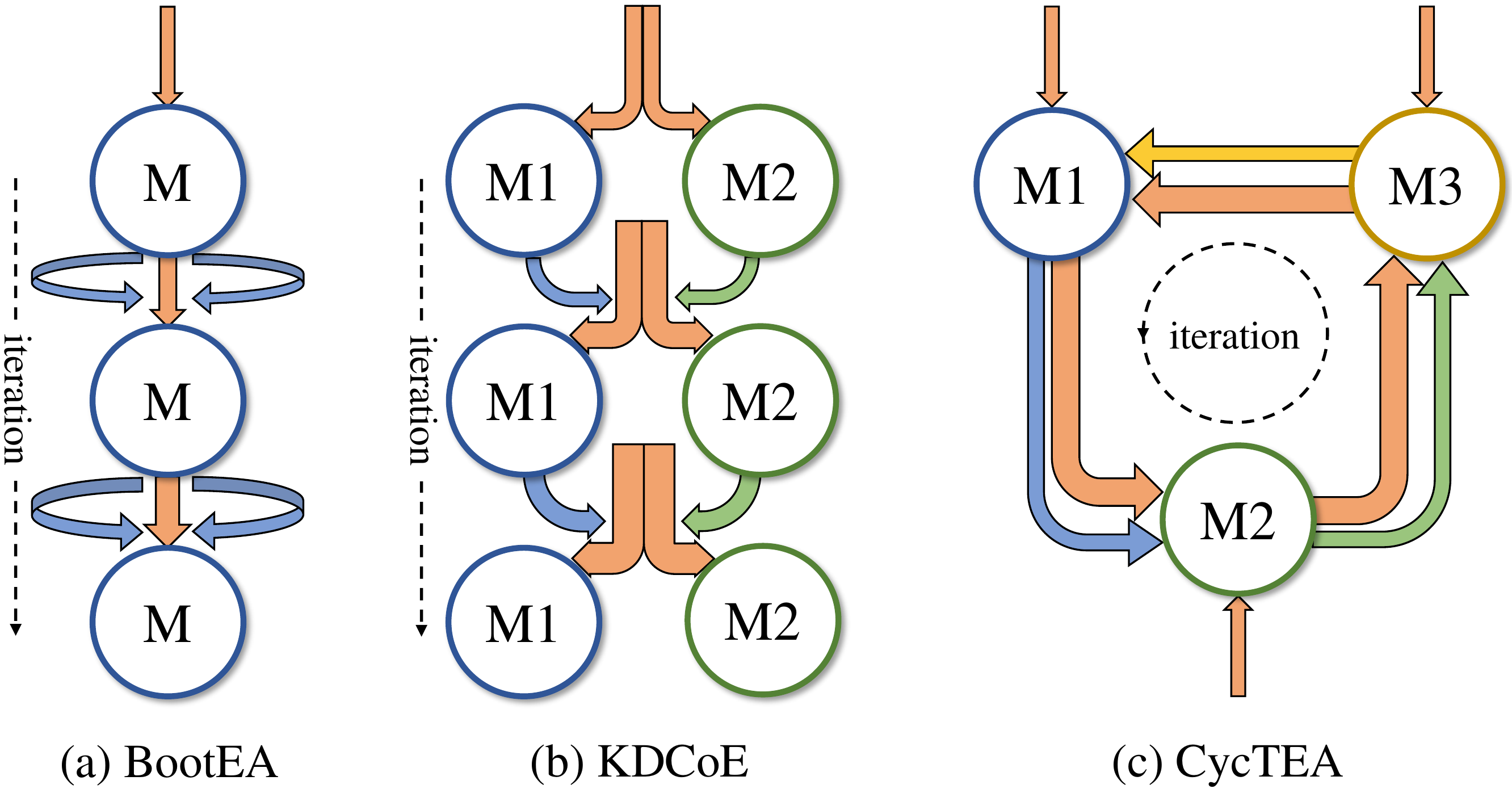}
    \caption{\label{fig:coteach.png}Comparison of BootEA, KDCoE, and the proposed \modelname. ``M'' stands for an aligner. Orange arrows represent training data. Blue, green and yellow arrows denote the error flows in the new alignment proposed by different aligners. 
    }
\end{figure}

\subsection{Discussions}

We compare \modelname with two semi-supervised approaches in Figure \ref{fig:coteach.png}. 
The self-training approach BootEA can be regarded as a special case of cycle-teaching with only one aligner. 
It directly feeds the selected entity alignment to itself. 
The noise is also transferred back to the aligner.
The co-training approach KDCoE utilizes two independent aligners to propose new alignment.
The noisy data from both aligners is also accumulated together. 
Therefore, it still suffers from the problem that each aligner's noisy information is also transferred back to itself. 
By contrast, in our framework, a large part of the aligner's noisy alignment is not directly sent back to itself due to the alignment cycle propagation. 
Instead, the noisy pairs are fed into the other model. 
As different aligners generate embeddings from different perspectives, one aligner's noisy pairs may be easily handled by the other. 
In addition, we carefully design the training data accumulation procedure as a fine-tuned step by removing the negative sampling from alignment learning over selected new pairs, and set smaller semi-supervised training epochs.
Therefore, the aligner will be adapted to the correct pairs from the other model firstly, and the influence of noisy data can be reduced.

\section{Experiment}
\label{sec:exp}
We build our framework on top of the OpenEA library \cite{sun2020benchmarking}. We will release our source code in GitHub\footnote{\url{https://github.com/JadeXIN/CycTEA}}. 

\subsection{Experimental Settings}
\textbf{Datasets.}
Current datasets such as DBP/DWY \cite{sun2017cross} are quite different from real-word KGs \cite{sun2020benchmarking}. Hence, we use the benchmark dataset released at the OpenEA library \cite{sun2020benchmarking} for evaluation, which follows the data distribution of real KGs. It contains two cross-lingual settings (English-to-French and English-to-German) and two monolingual settings (DBpedia-to-Wikidata and DBpedia-to-YAGO). Each setting has two sizes with 15K and 100K pairs of reference entity alignment, respectively. We follow the dataset splits in OpenEA, where 20\% reference alignment is for training, 10\% for validation and 70\% for test. 

\smallskip
\noindent\textbf{Implementation Details.}
\modelname can incorporate any number of aligners($k\geq2$). We choose three structure-based models as the aligners, i.e., AlignE, AliNet and RSN4EA. We follow their implementations settings used in OpenEA for a fair comparison. The order arrangement parameter $\epsilon = 0.2$. Performance is tested with 5-fold cross-validation to ensure an unbiased evaluation. Following the convention, we use Hits@1, Hits@5 and MRR as evaluation metrics, and higher scores of these metrics indicate better performance. 

\begin{table*}[!t]
	\centering
	\caption{Entity alignment results on 15K datasets. * means the results are produced by their source code.}
	\resizebox{0.8\textwidth}{!}{
		\begin{tabular}{lcclcclcclccl}
			\toprule
			\multirow{2}{*}{Methods} & \multicolumn{3}{c}{EN-FR-15K} & \multicolumn{3}{c}{EN-DE-15K} & \multicolumn{3}{c}{D-W-15K} & \multicolumn{3}{c}{D-Y-15K} \\
			\cmidrule(lr){2-4} \cmidrule(lr){5-7} \cmidrule(lr){8-10} \cmidrule(lr){11-13} 
			& Hits@1 & Hits@5 & MRR & Hits@1 & Hits@5 & MRR & Hits@1 & Hits@5 & MRR & Hits@1 & Hits@5 & MRR \\ \midrule
			MTransE & 0.247 & 0.467 & 0.351 & 0.307 & 0.518 & 0.407 & 0.259 & 0.461 & 0.354 & 0.463 & 0.675 & 0.559 \\
			AlignE & 0.357 & 0.611 & 0.473 & 0.552 & 0.741 & 0.638 & 0.406 & 0.627 & 0.506 & 0.551 & 0.743 & 0.636 \\
			BootEA & 0.507 & 0.718 & 0.603 & 0.675 & 0.820 & 0.740 & 0.572 & 0.744 & 0.649 & 0.739 & 0.849 & 0.788 \\
			SEA & 0.280 & 0.530 & 0.397 & 0.530 & 0.718 & 0.617 & 0.360 & 0.572 & 0.458 & 0.500 & 0.706 & 0.591 \\
			\midrule
			GCN-Align & 0.338 & 0.589 & 0.451 & 0.481 & 0.679 & 0.571 & 0.364 & 0.580 & 0.461 & 0.465 & 0.626 & 0.536 \\
			AliNet & 0.364 & 0.597 & 0.467 & 0.604 & 0.759 & 0.673 & 0.440 & 0.628 & 0.522 & 0.559 & 0.690 & 0.617 \\
			HyperKA* & 0.353 & 0.630 & 0.477 & 0.560 & 0.780 & 0.656 & 0.440 & 0.686 & 0.548 & 0.568 & 0.777 & 0.659 \\
			KE-GCN* & 0.408 & 0.670 & 0.524 & 0.658 & 0.822 & 0.730 & 0.519 & 0.727 & 0.608 & 0.560 & 0.750 & 0.644\\
			\midrule
			IPTransE & 0.169 & 0.320 & 0.243 & 0.350 & 0.515 & 0.430 & 0.232 & 0.380 & 0.303 & 0.313 & 0.456 & 0.378 \\
			RSN4EA & 0.393 & 0.595 & 0.487 & 0.587 & 0.752 & 0.662 & 0.441 & 0.615 & 0.521 & 0.514 & 0.655 & 0.580 \\
			\midrule
			AlignE+ & 0.563 & 0.765 & 0.653 & 0.707 & 0.859 & 0.775 & 0.633 & 0.798 & 0.706 & 0.742 & 0.854 & 0.791\\ 
			AliNet+ & 0.609 & 0.778 & 0.684 & 0.751 & 0.874 & 0.805 & 0.673 & 0.809 & 0.731 & \textbf{0.783} & 0.863 & 0.818\\
			RSN4EA+ & 0.524 & 0.721 & 0.612 & 0.697 & 0.846 & 0.762 & 0.595 & 0.746 & 0.663 & 0.670 & 0.770 & 0.715\\
			\midrule
			\modelname & \textbf{0.622} & \textbf{0.814} & \textbf{0.708} & \textbf{0.756} & \textbf{0.892} & \textbf{0.816} & \textbf{0.686} & \textbf{0.838} & \textbf{0.753} & 0.777 & \textbf{0.871} & \textbf{0.820} \\
			\bottomrule
	\end{tabular}}
	\label{tab:ent_alignment_15k}
\end{table*}

\begin{table*}[!t]
	\centering
	\caption{Entity alignment results on 100K datasets. * means the results are produced by their source code.}
	\resizebox{0.8\textwidth}{!}{
	\begin{tabular}{lcclcclcclccl}
		\toprule
		\multirow{2}{*}{Methods} & \multicolumn{3}{c}{EN-FR-100K} & \multicolumn{3}{c}{EN-DE-100K} & \multicolumn{3}{c}{D-W-100K} & \multicolumn{3}{c}{D-Y-100K} \\
		\cmidrule(lr){2-4} \cmidrule(lr){5-7} \cmidrule(lr){8-10} \cmidrule(lr){11-13} 
		& Hits@1 & Hits@5 & MRR & Hits@1 & Hits@5 & MRR & Hits@1 & Hits@5 & MRR & Hits@1 & Hits@5 & MRR \\
			\midrule
			MTransE & 0.138 & 0.261 & 0.202 & 0.140 & 0.264 & 0.204 & 0.210 & 0.358 & 0.282 & 0.244 & 0.414 & 0.328 \\
			AlignE & 0.294 & 0.483 & 0.388 & 0.423 & 0.593 & 0.505 & 0.385 & 0.587 & 0.478 & 0.617 & 0.776 & 0.691 \\
			BootEA & 0.389 & 0.561 & 0.474 & 0.518 & 0.673 & 0.592 & 0.516 & 0.685 & 0.594 & 0.703 & 0.827 & 0.761 \\
			SEA & 0.225 & 0.399 & 0.314 & 0.341 & 0.502 & 0.421 & 0.291 & 0.470 & 0.378 & 0.490 & 0.677 & 0.578 \\
			\midrule
			GCN-Align & 0.230 & 0.412 & 0.319 & 0.317 & 0.485 & 0.399 & 0.324 & 0.507 & 0.409 & 0.528 & 0.695 & 0.605 \\
			AliNet & 0.266 & 0.444 & 0.348 & 0.405 & 0.546 & 0.471 & 0.369 & 0.535 & 0.444 & 0.626 & 0.772 & 0.692 \\
			HyperKA* & 0.231 & 0.426 & 0.324 & 0.239 & 0.432 & 0.332 & 0.312 & 0.535 & 0.417 & 0.473 & 0.696 & 0.574\\
			KE-GCN* &  0.305 & 0.513 & 0.405 & 0.459 & 0.634 & 0.541 & 0.426 & 0.620 & 0.515 & 0.625 & 0.791 & 0.700 \\
			\midrule
			IPTransE & 0.158 & 0.277 & 0.219 & 0.226 & 0.357 & 0.292 & 0.221 & 0.352 & 0.285 & 0.396 & 0.558 & 0.474 \\
			RSN4EA & 0.293 & 0.452 & 0.371 & 0.430 & 0.570 & 0.497 & 0.384 & 0.533 & 0.454 & 0.620 & 0.769 & 0.688 \\
			\midrule
			AlignE+ & 0.379 & 0.558 & 0.466 & 0.509 & 0.659 & 0.581 & 0.492 & 0.665 & 0.571 & 0.699 & 0.831 & 0.759 \\
			AliNet+ & 0.429 & 0.600 & 0.510 & 0.546 & 0.676 & 0.608 & 0.555 & 0.708 & 0.623 & 0.742 & 0.864 & 0.796\\
			RSN4EA+ & 0.324 & 0.486 & 0.403 & 0.465 & 0.600 & 0.530 & 0.436 & 0.587 & 0.507 & 0.667 & 0.810 & 0.732\\
			\midrule
			\modelname & \textbf{0.442} & \textbf{0.627} & \textbf{0.530} & \textbf{0.560} & \textbf{0.709} & \textbf{0.630} & \textbf{0.566} & \textbf{0.732} & \textbf{0.641} & \textbf{0.747} & \textbf{0.871} & \textbf{0.803} \\
			\bottomrule
	\end{tabular}}
	\label{tab:ent_alignment_100k}
\end{table*}

\smallskip
\noindent\textbf{Baselines.}
Similarly, we compare with structure-based entity alignment models for a fair comparison:
\begin{itemize}
\item Triple-based models that capture the local semantics information of relation triples based on TransE, including MTransE \cite{chen2017multilingual}, AlignE and BootEA \cite{sun2018bootstrapping} as well as SEA \cite{pei2019semi}.
\item Neighborhood-based models that use GNNs to exploit subgraph structures for entity alignment, including GCNAlign \cite{wang2018cross}, AliNet \cite{sun2020knowledge}, HyperKA \cite{sun2020knowledge} and KE-GCN \cite{yu2021knowledge}.
\item Path-based models that explore the long-term dependency across relation paths, including IPTransE \cite{zhu2017iterative} and RSN4EA\cite{guo2018recurrent}.
\end{itemize}

We do not compare with some recent attribute-based models \cite{chen2018co,zhang2019multi,wu2019relation,zhang2019multi} since they require side information that our framework, as well as other baselines, do not use. In addition, as RREA \cite{mao2020relational} failed in OpenEA 100K datasets \cite{ge2021largeea}, we do not include it in the baselines.

\subsection{Main Results}
Tables \ref{tab:ent_alignment_15k} and \ref{tab:ent_alignment_100k} present the entity alignment results.
\modelname outperforms all baselines, and is 4\% to 11\% higher than the strongest baseline BootEA on Hits@1. e.g., \modelname outperforms BootEA by 11.5\% on EN-FR-15K. BootEA achieves second-best results due to its bootstrapping strategy, but the limited ability of self-training prevents its further improvement. In the supervised setting, KE-GCN, AliNet and RSN4EA all acquire satisfactory results, but lack of training data limits their performance. On 100K datasets, many baselines fail to achieve promising results due to more complex KG structures and larger alignment space, but \modelname maintains the best performance, demonstrating its practicability. 
The variant of an aligner in \modelname is denoted as ``aligner+'' (e.g., ``AlignE+'' refers to the ``AlignE'' in \modelname). Three aligners all acquire large improvement in \modelname. For example, AliNet benefits the most from cycle-teaching because it captures two-hop information and the high-quality new alignment can boost it largely. 
The result of \modelname is a little lower than AliNet+ in D-Y-15K because the large performance variance of three aligners hurts the ensemble result.

\subsection{Further Analyses}
\smallskip
\noindent\textbf{Effectiveness of Cycle-teaching.}
Table \ref{tab:coteach-effect} compares \modelname with other semi-supervised strategies. The first three model variants use self-training. \modelname significantly outperforms the self-training models, because it can integrate knowledge from all aligners. Compared with other alignment selection strategies, i.e., keeping the alignment supported by all aligners (Intersection), or combining alignment from all aligners (Union), or selecting alignment supported by the highest number of aligners (Majority vote), \modelname still achieves the best performance. The improvement achieved by ``Intersection'' is the smallest because entity pairs supported by all aligners are limited. But it has the best result in EN-FR by coincidence, as the noisy alignment proposed by all aligners is higher than other datasets, and ``Intersection" can filter the noise thoroughly. ''Union'' achieves limited improvement since it involves much noise. ``Majority vote'' is more considerable than them but still lower than \modelname as it inevitably removes some correct entity alignment pairs.

\begin{table}[!t]
\centering
\caption{Hits@1 scores w.r.t. different semi-supervised learning strategies on the 15K datasets.}
\resizebox{0.8\linewidth}{!}{
\begin{tabular}{lcccc} 
\toprule
Methods & EN-FR & EN-DE & D-W & D-Y \\
\midrule
AlignE (semi) & 0.507 & 0.675 & 0.572 & 0.739 \\ 
RSN4EA (semi) & 0.497 & 0.673 & 0.553 & 0.632 \\
AliNet (semi) & 0.452 & 0.685 & 0.570 & 0.753 \\ 
\midrule
\modelname (Intersection) & \textbf{0.632} & 0.726 & 0.631 & 0.752\\
\modelname (Union) & 0.594 & 0.736 & 0.657 & 0.750\\
\modelname (Majority vote) & 0.614 & 0.739 & 0.677 & 0.761\\
\midrule
\modelname & 0.622 & \textbf{0.756} & \textbf{0.686} & \textbf{0.777} \\ 
\bottomrule
\end{tabular}}
\label{tab:coteach-effect}
\end{table}

\smallskip
\noindent\textbf{Effectiveness of Re-matching.}
Table \ref{tab:acc_align} evaluates the effect of our re-matching strategy for conflict resolution (cr). We can see that re-matching can significantly improve the precision of new alignment by $4\%$ to $9\%$, while with a slight decrease in recall. This is because the filter process of re-matching inevitably breaks some correct entity pairs. Finally, re-matching can bring $1\%$ to $3\%$ improvement on F1-score to all aligners and further improve the overall performance.

\begin{table}[!t]
	\centering
	\caption{Quality of proposed new alignment on 15K datasets}
	\resizebox{0.9\linewidth}{!}{
		\begin{tabular}{lcclccl}
			\toprule
			\multirow{2}{*}{Methods} & \multicolumn{3}{c}{EN-FR-15K} & \multicolumn{3}{c}{D-W-15K} \\
			\cmidrule(lr){2-4} \cmidrule(lr){5-7} 
			& Prec. & Rec. & F1 & Prec. & Rec. & F1 \\ \midrule
			AlignE+ (w/o cr) & 0.650 & 0.569 & 0.606 & 0.718 & 0.639 & 0.676 \\
			AlignE+ & 0.696 & 0.541 & 0.609 & 0.761 & 0.622 & 0.685 \\
			\midrule
			RSN4EA+ (w/o cr) & 0.619 & 0.538 & 0.575 & 0.649 & 0.607 & 0.627\\
			RSN4EA+ & 0.684 & 0.528 & 0.596 & 0.717 & 0.602 & 0.655\\
			\midrule
			AliNet+ (w/o cr) & 0.601 & 0.560 & 0.579 & 0.675 & 0.640 & 0.657 \\
			AliNet+ & 0.687 & 0.534 & 0.601 & 0.766 & 0.608 & 0.678 \\
			\bottomrule
	\end{tabular}}
	\label{tab:acc_align}
\end{table}

\begin{figure}[!t]
    \centering
    \includegraphics[width=0.85\linewidth]{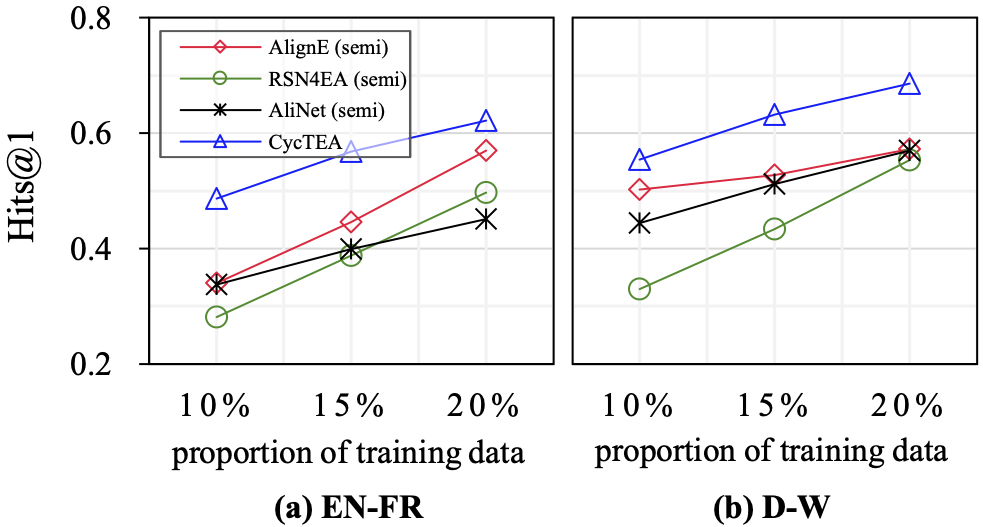}
    \caption{\label{fig:train_ratio}Hits@1 w.r.t different proportions of training data.}
\end{figure}

\smallskip
\noindent\textbf{Robustness to Noisy Accumulated Entity Alignment.}
To evaluate the robustness of \modelname to the accumulated noisy alignment, we evaluate the framework trained on different proportions of seed entity alignment from 10\% to 20\% to simulate this scenario, as less training data leads to worse performance and a larger ratio of noise in the proposed new alignment. Figure \ref{fig:train_ratio} depicts the trend of Hits@1 of three aligners in self-training (denoted as ``aligner (semi)''). They all achieve better performance as the ratio increases. We can see AliNet (semi) is more sensitive to the training data ratio as its performance drops drastically when the training data size decreases. As GNNs capture the global structure information, the useful information it can capture reduces exponentially when the training data is meager. RSN4EA (semi) has poor robustness as it presents much worse results in the scenarios of heavy noise. \modelname maintains promising performance over any training data ratio, which validates its superiority. 

\begin{figure}[!t]
    \centering
    \includegraphics[width=0.95\linewidth]{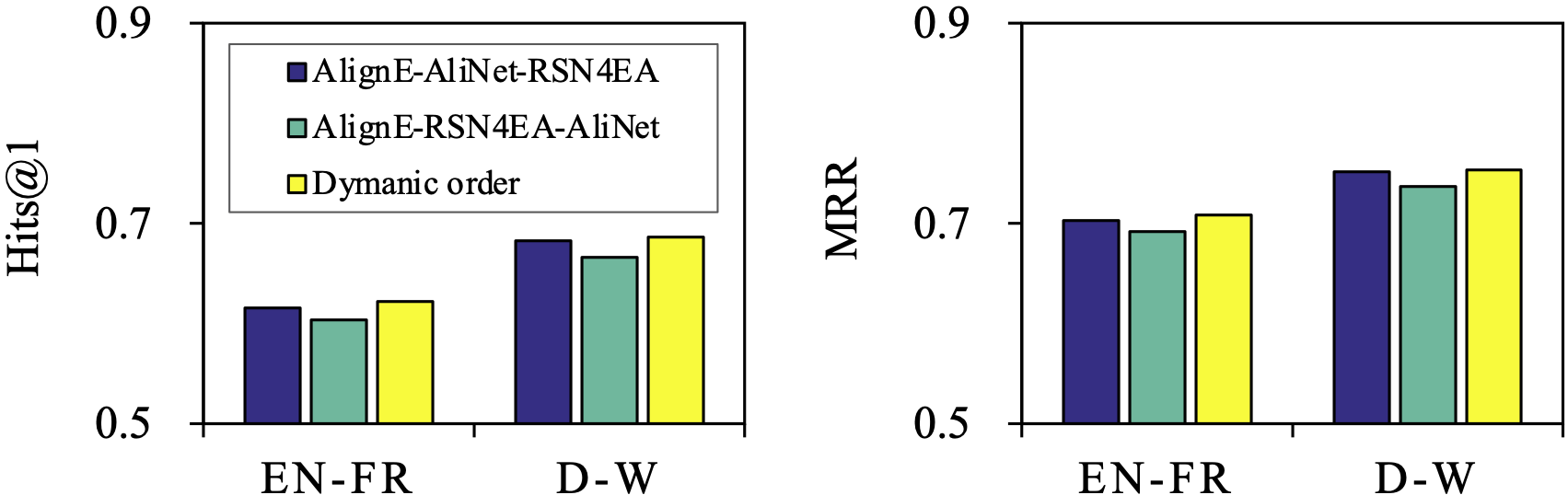}
    \caption{\label{fig:order}Hits@1 and MRR w.r.t different teaching orders.}
\end{figure}

\smallskip
\noindent\textbf{Effectiveness of Dynamic Order Arrangement.}
Figure \ref{fig:order} represents the performance with different cycle-teaching orders.
There are two static orders in total given three aligners. We can see the first order is superior to the second one, and our dynamic order exceeds both static orders. In particular, the performance with the order ``AlignE-AliNet-RSN4EA'' is only slightly worse than ours. This is because the order appears much more frequent than the other one in our dynamic order arrangement during iteration. This shows that our dynamic order arrangement can effectively capture the optimal order during training and result in better performance.

\smallskip
\noindent\textbf{Effectiveness of Diversity-aware Alignment Selection.}
Table~\ref{tab:diversity} gives the ablation study on our diversity-aware selection method. We report the results of AliNet in \modelname due to space limitation, where ``w/o daas'' denotes the variant without using the matching diversity for alignment selection. We can see that our diversity-aware method can improve the precision and F1-score of selected alignment and reduce noise.

\begin{table}[!t]
	\centering
	\caption{Quality of the new alignment proposed by AliNet in \modelname w/ and w/o diversity-aware alignment selection.}
	\resizebox{0.9\linewidth}{!}{
		\begin{tabular}{lcclccl}
			\toprule
			\multirow{2}{*}{Methods} & \multicolumn{3}{c}{EN-FR-15K} & \multicolumn{3}{c}{D-W-15K} \\
			\cmidrule(lr){2-4} \cmidrule(lr){5-7} 
			& Prec. & Rec. & F1 & Prec. & Rec. & F1 \\ \midrule
			 AliNet+(w/o daas) & 0.591 & 0.565 & 0.578 & 0.672 & 0.652 & 0.661 \\
			 AliNet+ & 0.687 & 0.534 & 0.601 & 0.766 & 0.608 & 0.678 \\
			\bottomrule
	\end{tabular}}
	\label{tab:diversity}
\end{table}

\begin{figure}[!t]
    \centering
    \includegraphics[width=0.75\linewidth]{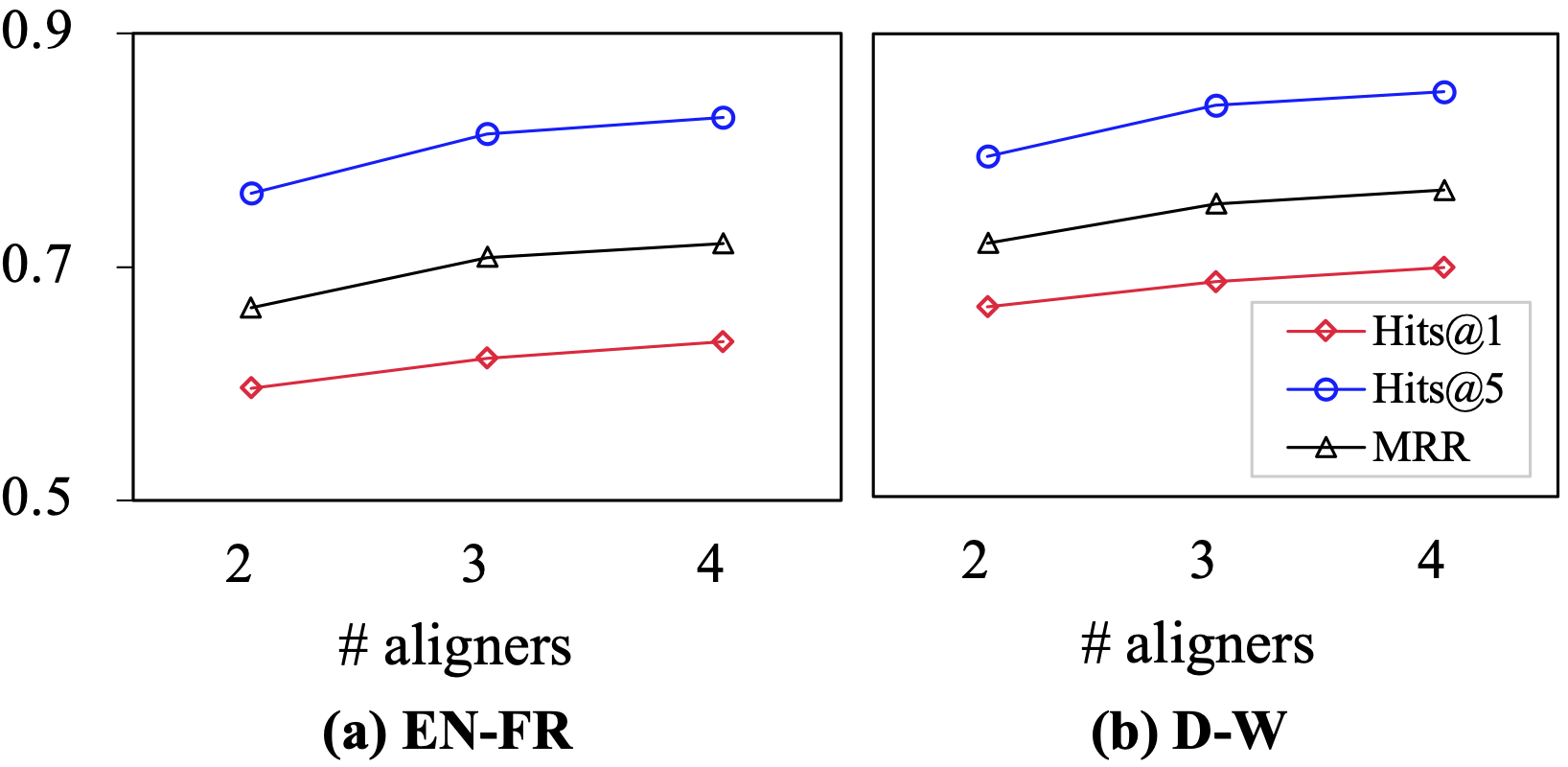}
    \caption{\label{fig:multi-aligner}Entity alignment performance w.r.t. \# aligners.}
\end{figure}

\smallskip
\noindent\textbf{Different numbers of aligners in \modelname.}
Our framework can be implemented to any number of aligners.
Due to space limitation, we test its performance with $k=2,3,4$ aligners, respectively. 
We choose AlignE and AliNet for $k=2$, and add RSN4EA for $k=3$, and introduce KE-GCN for $k=4$. These aligners are well-performing structure-based models, and these combinations are the optimal settings given these four aligners. 
Figure \ref{fig:multi-aligner} indicates that the performance increases when integrating more aligners. 
But in the main setting, we choose AlignE, AliNet and RSN4EA, the structure-based aligners with high complementarity and good performance, for a trade-off between effectiveness and efficiency.

\section{Conclusion}
\label{sec:conclusion}
We present a novel and practical cycle-teaching framework for entity alignment. It utilizes multiple entity alignment models and iteratively cycle-teach each model by transmitting their proposed reliable alignment. Cycle-teaching can primarily remedy the effect of noisy data in the accumulated new alignment and extend all models' alignment learning ability. Our diversity-aware alignment selection and re-matching based alignment conflict resolution strategies further improve the quality of the new alignment. We also consider the effect of teaching order and propose dynamic order arrangement. Experiments on benchmark datasets show that our approach can outperform SOTA methods and achieve promising results in the heavy noise-propagation scenario. 
For future work, we plan to extend our framework to multi-view features.

\section{Acknowledgments}
This work was partially supported by the Australian Research Council under Grant No. DE210100160.
Zequn Sun's work was supported by Program A for Outstanding PhD Candidates of Nanjing University.

\bibliography{aaai22}

\begin{thebibliography}{36}
\providecommand{\natexlab}[1]{#1}

\bibitem[{Bordes et~al.(2013)Bordes, Usunier, Garc{\'{\i}}a{-}Dur{\'{a}}n,
  Weston, and Yakhnenko}]{bordes2013translating}
Bordes, A.; Usunier, N.; Garc{\'{\i}}a{-}Dur{\'{a}}n, A.; Weston, J.; and
  Yakhnenko, O. 2013.
\newblock Translating Embeddings for Modeling Multi-relational Data.
\newblock In \emph{Proceedings of the 27th Annual Conference on Neural
  Information Processing Systems}, 2787--2795.

\bibitem[{Cao et~al.(2019)Cao, Liu, Li, Li, and Chua}]{cao2019multi}
Cao, Y.; Liu, Z.; Li, C.; Li, J.; and Chua, T.-S. 2019.
\newblock Multi-Channel Graph Neural Network for Entity Alignment.
\newblock In \emph{Proceedings of the 57th Conference of the Association for
  Computational Linguistics}, 1452--1461.

\bibitem[{Chen et~al.(2018)Chen, Tian, Chang, Skiena, and Zaniolo}]{chen2018co}
Chen, M.; Tian, Y.; Chang, K.-W.; Skiena, S.; and Zaniolo, C. 2018.
\newblock Co-training Embeddings of Knowledge Graphs and Entity Descriptions
  for Cross-lingual Entity Alignment.
\newblock In \emph{Proceedings of the 27th International Joint Conference on
  Artificial Intelligence}, 3998--4004.

\bibitem[{Chen et~al.(2017)Chen, Tian, Yang, and
  Zaniolo}]{chen2017multilingual}
Chen, M.; Tian, Y.; Yang, M.; and Zaniolo, C. 2017.
\newblock Multilingual knowledge graph embeddings for cross-lingual knowledge
  alignment.
\newblock In \emph{Proceedings of the 26th International Joint Conference on
  Artificial Intelligence}, 1511--1517.

\bibitem[{Fey et~al.(2020)Fey, Lenssen, Morris, Masci, and
  Kriege}]{fey2020deep}
Fey, M.; Lenssen, J.~E.; Morris, C.; Masci, J.; and Kriege, N.~M. 2020.
\newblock Deep Graph Matching Consensus.
\newblock In \emph{Proceedings of 8th International Conference on Learning
  Representations}.

\bibitem[{Gal, Roitman, and Sagi(2016)}]{mcd_www16}
Gal, A.; Roitman, H.; and Sagi, T. 2016.
\newblock From Diversity-based Prediction to Better Ontology {\&} Schema
  Matching.
\newblock In \emph{Proceedings of the 25th International Conference on World
  Wide Web}, 1145--1155.

\bibitem[{Ge et~al.(2021)Ge, Liu, Chen, Zheng, and Gao}]{ge2021largeea}
Ge, C.; Liu, X.; Chen, L.; Zheng, B.; and Gao, Y. 2021.
\newblock LargeEA: Aligning Entities for Large-scale Knowledge Graphs.
\newblock \emph{arXiv preprint arXiv:2108.05211}.

\bibitem[{Guo, Sun, and Hu(2019)}]{guo2018recurrent}
Guo, L.; Sun, Z.; and Hu, W. 2019.
\newblock Learning to Exploit Long-term Relational Dependencies in Knowledge
  Graphs.
\newblock In \emph{Proceedings of the 36th International Conference on Machine
  Learning}, 2505--2514.

\bibitem[{Han et~al.(2018)Han, Yao, Yu, Niu, Xu, Hu, Tsang, and
  Sugiyama}]{han2018coteach}
Han, B.; Yao, Q.; Yu, X.; Niu, G.; Xu, M.; Hu, W.; Tsang, I.~W.; and Sugiyama,
  M. 2018.
\newblock Co-teaching: Robust Training of Deep Neural Networks with Extremely
  Noisy Labels.
\newblock In \emph{Proceedings of the Annual Conference on Neural Information
  Processing Systems}, 8536--8546.

\bibitem[{Kipf and Welling(2017)}]{kipf2017gcn}
Kipf, T.~N.; and Welling, M. 2017.
\newblock Semi-Supervised Classification with Graph Convolutional Networks.
\newblock In \emph{Proceedings of the 5th International Conference on Learning
  Representations}.

\bibitem[{Li et~al.(2019)Li, Cao, Hou, Shi, Li, and Chua}]{li2019semi}
Li, C.; Cao, Y.; Hou, L.; Shi, J.; Li, J.; and Chua, T. 2019.
\newblock Semi-supervised Entity Alignment via Joint Knowledge Embedding Model
  and Cross-graph Model.
\newblock In \emph{Proceedings of the 2019 Conference on Empirical Methods in
  Natural Language Processing and the 9th International Joint Conference on
  Natural Language Processing}, 2723--2732.

\bibitem[{Liu et~al.(2020)Liu, Cao, Pan, Li, and Chua}]{liu2020exploring}
Liu, Z.; Cao, Y.; Pan, L.; Li, J.; and Chua, T.-S. 2020.
\newblock Exploring and Evaluating Attributes, Values, and Structure for Entity
  Alignment.
\newblock In \emph{Proceedings of the 2020 Conference on Empirical Methods in
  Natural Language Processing}, 6355--6364.

\bibitem[{Mao et~al.(2020)Mao, Wang, Xu, Wu, and Lan}]{mao2020relational}
Mao, X.; Wang, W.; Xu, H.; Wu, Y.; and Lan, M. 2020.
\newblock Relational Reflection Entity Alignment.
\newblock In \emph{Proceedings of the 29th {ACM} International Conference on
  Information and Knowledge Management}, 1095--1104.

\bibitem[{Pei et~al.(2019)Pei, Yu, Hoehndorf, and Zhang}]{pei2019semi}
Pei, S.; Yu, L.; Hoehndorf, R.; and Zhang, X. 2019.
\newblock Semi-supervised Entity Alignment via Knowledge Graph Embedding with
  Awareness of Degree Difference.
\newblock In \emph{Proceedings of the World Wide Web Conference}, 3130--3136.

\bibitem[{Suchanek, Abiteboul, and Senellart(2011)}]{suchanek2011paris}
Suchanek, F.~M.; Abiteboul, S.; and Senellart, P. 2011.
\newblock {PARIS:} Probabilistic Alignment of Relations, Instances, and Schema.
\newblock \emph{Proceedings of the VLDB Endowment}, 5(3): 157--168.

\bibitem[{Sun, Hu, and Li(2017)}]{sun2017cross}
Sun, Z.; Hu, W.; and Li, C. 2017.
\newblock Cross-lingual Entity Alignment via Joint Attribute-Preserving
  Embedding.
\newblock In \emph{Proceedings of the 16th International Semantic Web
  Conference}, 628--644.

\bibitem[{Sun et~al.(2018)Sun, Hu, Zhang, and Qu}]{sun2018bootstrapping}
Sun, Z.; Hu, W.; Zhang, Q.; and Qu, Y. 2018.
\newblock Bootstrapping Entity Alignment with Knowledge Graph Embedding.
\newblock In \emph{Proceedings of the 27th International Joint Conference on
  Artificial Intelligence}, 4396--4402.

\bibitem[{Sun et~al.(2020{\natexlab{a}})Sun, Wang, Hu, Chen, Dai, Zhang, and
  Qu}]{sun2020knowledge}
Sun, Z.; Wang, C.; Hu, W.; Chen, M.; Dai, J.; Zhang, W.; and Qu, Y.
  2020{\natexlab{a}}.
\newblock Knowledge Graph Alignment Network with Gated Multi-hop Neighborhood
  Aggregation.
\newblock \emph{Proceedings of the 34th {AAAI} Conference on Artificial
  Intelligence}, 222--229.

\bibitem[{Sun et~al.(2020{\natexlab{b}})Sun, Zhang, Hu, Wang, Chen, Akrami, and
  Li}]{sun2020benchmarking}
Sun, Z.; Zhang, Q.; Hu, W.; Wang, C.; Chen, M.; Akrami, F.; and Li, C.
  2020{\natexlab{b}}.
\newblock A Benchmarking Study of Embedding-based Entity Alignment for
  Knowledge Graphs.
\newblock \emph{Proceedings of the VLDB Endowment}, 13(11): 2326--2340.

\bibitem[{Tang et~al.(2020)Tang, Zhang, Chen, Yang, Chen, and
  Li}]{tang2020bertint}
Tang, X.; Zhang, J.; Chen, B.; Yang, Y.; Chen, H.; and Li, C. 2020.
\newblock {BERT-INT:} {A} BERT-based Interaction Model For Knowledge Graph
  Alignment.
\newblock In \emph{Proceedings of the 29th International Joint Conference on
  Artificial Intelligence}, 3174--3180.

\bibitem[{Trisedya, Qi, and Zhang(2019)}]{trisedya2019entity}
Trisedya, B.~D.; Qi, J.; and Zhang, R. 2019.
\newblock Entity Alignment between Knowledge Graphs Using Attribute Embeddings.
\newblock In \emph{Proceedings of the 33rd {AAAI} Conference on Artificial
  Intelligence}, 297--304.

\bibitem[{Wang et~al.(2017)Wang, Mao, Wang, and Guo}]{wang2017knowledge}
Wang, Q.; Mao, Z.; Wang, B.; and Guo, L. 2017.
\newblock Knowledge Graph Embedding: {A} Survey of Approaches and Applications.
\newblock \emph{IEEE Transactions on Knowledge and Data Engineering}, 29(12):
  2724--2743.

\bibitem[{Wang et~al.(2018)Wang, Lv, Lan, and Zhang}]{wang2018cross}
Wang, Z.; Lv, Q.; Lan, X.; and Zhang, Y. 2018.
\newblock Cross-lingual Knowledge Graph Alignment via Graph Convolutional
  Networks.
\newblock In \emph{Proceedings of the 2018 Conference on Empirical Methods in
  Natural Language Processing}, 349--357.

\bibitem[{Wu et~al.(2019{\natexlab{a}})Wu, Liu, Feng, Wang, Yan, and
  Zhao}]{wu2019relation}
Wu, Y.; Liu, X.; Feng, Y.; Wang, Z.; Yan, R.; and Zhao, D. 2019{\natexlab{a}}.
\newblock Relation-Aware Entity Alignment for Heterogeneous Knowledge Graphs.
\newblock In \emph{Proceedings of the 28th International Joint Conference on
  Artificial Intelligence}, 5278--5284.

\bibitem[{Wu et~al.(2019{\natexlab{b}})Wu, Liu, Feng, Wang, and
  Zhao}]{wu2019jointly}
Wu, Y.; Liu, X.; Feng, Y.; Wang, Z.; and Zhao, D. 2019{\natexlab{b}}.
\newblock Jointly Learning Entity and Relation Representations for Entity
  Alignment.
\newblock In \emph{Proceedings of the 2019 Conference on Empirical Methods in
  Natural Language Processing and the 9th International Joint Conference on
  Natural Language Processing}, 240--249.

\bibitem[{Wu et~al.(2020)Wu, Liu, Feng, Wang, and Zhao}]{wu2020neighborhood}
Wu, Y.; Liu, X.; Feng, Y.; Wang, Z.; and Zhao, D. 2020.
\newblock Neighborhood Matching Network for Entity Alignment.
\newblock In \emph{Proceedings of the 58th Annual Meeting of the Association
  for Computational Linguistics}, 6477--6487.

\bibitem[{Xu et~al.(2019)Xu, Wang, Yu, Feng, Song, Wang, and Yu}]{xu2019cross}
Xu, K.; Wang, L.; Yu, M.; Feng, Y.; Song, Y.; Wang, Z.; and Yu, D. 2019.
\newblock Cross-lingual Knowledge Graph Alignment via Graph Matching Neural
  Network.
\newblock \emph{Proceedings of the 57th Conference of the Association for
  Computational Linguistics}, 3156--3161.

\bibitem[{Yang et~al.(2019)Yang, Zou, Shi, Lu, Lin, and Sun}]{yang2019aligning}
Yang, H.; Zou, Y.; Shi, P.; Lu, W.; Lin, J.; and Sun, X. 2019.
\newblock Aligning Cross-Lingual Entities with Multi-Aspect Information.
\newblock In \emph{Proceedings of the 2019 Conference on Empirical Methods in
  Natural Language Processing and the 9th International Joint Conference on
  Natural Language Processing}, 4430--4440.

\bibitem[{Yang et~al.(2020)Yang, Liu, Zhao, Wang, and Xie}]{COTSAE}
Yang, K.; Liu, S.; Zhao, J.; Wang, Y.; and Xie, B. 2020.
\newblock {COTSAE:} CO-Training of Structure and Attribute Embeddings for
  Entity Alignment.
\newblock In \emph{Proceedings of the 34th {AAAI} Conference on Artificial
  Intelligence}, 3025--3032.

\bibitem[{Ye et~al.(2019)Ye, Li, Fang, Zang, and Wang}]{ye2019vectorized}
Ye, R.; Li, X.; Fang, Y.; Zang, H.; and Wang, M. 2019.
\newblock A Vectorized Relational Graph Convolutional Network for
  Multi-Relational Network Alignment.
\newblock In \emph{Proceedings of the 28th International Joint Conference on
  Artificial Intelligence}, 4135--4141.

\bibitem[{Yu et~al.(2021)Yu, Yang, Zhang, and Wu}]{yu2021knowledge}
Yu, D.; Yang, Y.; Zhang, R.; and Wu, Y. 2021.
\newblock Knowledge Embedding Based Graph Convolutional Network.
\newblock In \emph{Proceedings of the Web Conference 2021}, 1619--1628.

\bibitem[{Zeng et~al.(2020)Zeng, Zhao, Wang, Tang, and Tan}]{zeng2020degree}
Zeng, W.; Zhao, X.; Wang, W.; Tang, J.; and Tan, Z. 2020.
\newblock Degree-Aware Alignment for Entities in Tail.
\newblock In \emph{Proceedings of the 43rd International {ACM} {SIGIR}
  Conference on Research and Development in Information Retrieval}, 811--820.

\bibitem[{Zhang et~al.(2019)Zhang, Sun, Hu, Chen, Guo, and Qu}]{zhang2019multi}
Zhang, Q.; Sun, Z.; Hu, W.; Chen, M.; Guo, L.; and Qu, Y. 2019.
\newblock Multi-view Knowledge Graph Embedding for Entity Alignment.
\newblock In \emph{Proceedings of the 28th International Joint Conference on
  Artificial Intelligence}, 5429--5435.

\bibitem[{Zhou and Li(2005)}]{Zhou05tritraining}
Zhou, Z.; and Li, M. 2005.
\newblock Tri-Training: Exploiting Unlabeled Data Using Three Classifiers.
\newblock \emph{IEEE Transactions on Knowledge and Data Engineering}, 17(11):
  1529--1541.

\bibitem[{Zhu et~al.(2017)Zhu, Xie, Liu, and Sun}]{zhu2017iterative}
Zhu, H.; Xie, R.; Liu, Z.; and Sun, M. 2017.
\newblock Iterative Entity Alignment via Joint Knowledge Embeddings.
\newblock In \emph{Proceedings of the 26th International Joint Conference on
  Artificial Intelligence}, 4258--4264.

\bibitem[{Zhu et~al.(2019)Zhu, Zhou, Wu, Tan, and Guo}]{zhu2019neighborhood}
Zhu, Q.; Zhou, X.; Wu, J.; Tan, J.; and Guo, L. 2019.
\newblock Neighborhood-Aware Attentional Representation for Multilingual
  Knowledge Graphs.
\newblock In \emph{Proceedings of the 28th International Joint Conference on
  Artificial Intelligence}, 1943--1949.

\end{thebibliography}

\end{document}